\documentclass[conference]{IEEEtran}
\IEEEoverridecommandlockouts
\usepackage{cite}
\usepackage{amsmath,amssymb,amsfonts}
\usepackage{algorithmic}
\usepackage{graphicx}
\usepackage{textcomp}
\usepackage{xcolor}
\usepackage{multirow}
\usepackage{booktabs}
\usepackage[linesnumbered,ruled,vlined]{algorithm2e}
\usepackage{threeparttable}
\def\BibTeX{{\rm B\kern-.05em{\sc i\kern-.025em b}\kern-.08em
    T\kern-.1667em\lower.7ex\hbox{E}\kern-.125emX}}
\begin{document}

\title{Efficient Traffic Forecasting on Large-Scale Road Network by Regularized Adaptive Graph Convolution}


\author{
    Kaiqi Wu$^{1}$, Weiyang Kong$^{1}$, Sen Zhang$^1$, Zitong Chen$^{2}$, Yubao Liu$^{1, 3}$\\
    \vspace{0.2em}
    \small
    $^1$School of Computer Science and Engineering, Sun Yat-Sen University, China \\
    $^2$School of Artificial Intelligence, Sun Yat-Sen University, China \\
    $^3$Guangdong Key Laboratory of Big Data Analysis and Processing, China \\
    \vspace{0.2em}
    \{wukq5, kongwy3, zhangs7\}@mail2.sysu.edu.cn, \{chenzt53, liuyubao\}@mail.sysu.edu.cn 
}

\maketitle

\begin{abstract}
Traffic prediction is a critical task in spatial-temporal forecasting with broad applications in travel planning and urban management. To model the complex spatial-temporal dependencies in traffic data, Spatial-Temporal Graph Convolutional Networks (STGCNs) have been widely employed, achieving advanced performance. However, when applied to large-scale road networks, the quadratic computational complexity of traditional graph convolution operations severely limits their scalability. Several methods attempt to address this issue through approximation, compression, or spatial partitioning. Nevertheless, these methods often either fail to achieve sufficient computational efficiency or compromise prediction accuracy. To address these challenges, we propose a Regularized Adaptive Graph Convolution (RAGC) model. First, to ensure scalability on large road networks, we develop the Efficient Cosine Operator (ECO), which performs graph convolution based on the cosine similarity of node embeddings with linear time complexity. Second, we introduce a regularized adaptive graph convolution framework that combines Stochastic Shared Embedding (SSE) and adaptive graph convolution through a residual difference mechanism. This design enables the model to learn high-quality node embeddings, thereby improving prediction accuracy while maintaining computational efficiency. Extensive experiments on four large-scale real-world traffic datasets show that RAGC consistently outperforms state-of-the-art methods in terms of prediction accuracy and exhibits competitive computational efficiency. The code is available at: https://github.com/wkq-wukaiqi/RAGC.
\end{abstract}


\section{Introduction}
Traffic forecasting aims to forecast future traffic conditions based on historical traffic data observed by traffic network sensors \cite{yin2021survey}. It has widespread applications in real life, including travel planning, congestion control, and assisting urban planning \cite{DGCRN, survey2, One4AllST, ST-ABC}.\par
In early studies, traditional statistical methods such as Support Vector Regression (SVR) \cite{SVR}, Historical Average (HA) \cite{HA}, and Autoregressive Integrated Moving Average (ARIMA) \cite{ARIMA} were widely applied to traffic flow prediction. However, these approaches fail to consider the spatial dependencies among nodes in a road network, rendering them unsuitable for capturing the complexities of real-world traffic data. In recent years, Spatial-Temporal Graph Neural Networks (STGNN) have emerged as the dominant framework for traffic forecasting due to their superior ability to model both spatial and temporal dependencies \cite{STGCN, DCRNN, GWNet, AGCRN}.\par

\begin{figure}[tp]
  \centering
  \includegraphics[width=0.45\textwidth]{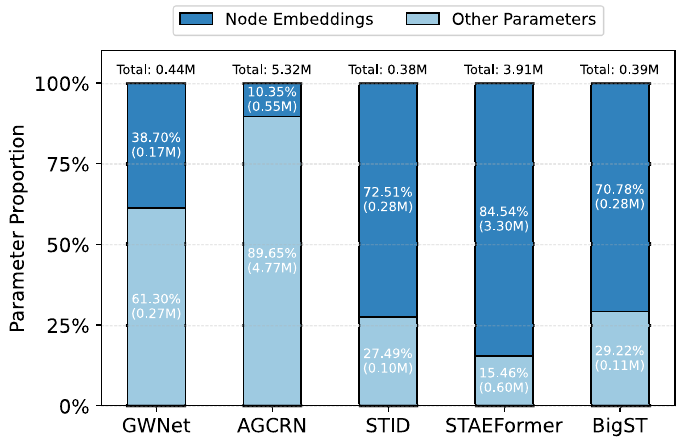}
  \caption{Parameter proportion analysis of different models on LargeST-CA (8,600 nodes), where 1M = 1,000,000.}
  \label{fig:intro}
\end{figure}

Early STGNN models relied on predefined adjacency matrices, either based on spatial distances between nodes \cite{STGCN, DCRNN} or on historical sequence similarity \cite{STFGNN, STGODE}. However, these static graph structures may not accurately reflect latent adjacency relationships between nodes. To address this limitation, adaptive spatial-temporal graph neural networks have been proposed, achieving state-of-the-art performance and gaining widespread adoption \cite{GWNet, AGCRN, MTGNN, HimNet}. The core idea behind adaptive graph learning is to introduce learnable node embeddings, enabling the model to capture node-level characteristics in a data-driven manner. \par
Regarding the use of node embeddings, existing methods can be broadly categorized into two branches. One branch utilizes node embeddings to construct adaptive adjacency matrices for use in graph convolution operations, as seen in models like GWNet \cite{GWNet}, MTGNN \cite{MTGNN}, and AGCRN \cite{AGCRN}. The other branch does not explicitly construct adjacency matrices. Instead, it concatenates node embeddings with input sequences along the feature dimension to create enriched representations, as implemented in STID \cite{STID} and STAEFormer \cite{STAEFormer}. While these approaches have demonstrated excellent prediction performance, there are still some limitations.\par
First, in the adaptive adjacency matrix based methods, although node embeddings eliminate the need to explicitly learn a full adjacency matrix and thereby reduce the number of training parameters, adaptive graph convolution still incurs a computational complexity of $O(N^2)$, where $N$ denotes the number of nodes in the road network. This quadratic complexity significantly limits the scalability of STGNN models in large-scale traffic networks. To enhance scalability, models like BigST \cite{BigST} have reduced graph convolution complexity to $O(N)$. However, this is achieved by using random feature maps to approximate the softmax kernel, a strategy that introduces additional noise \cite{Performer} and may adversely affect both training stability and final model performance. PatchSTG \cite{PatchSTG} reduces complexity via a KD-tree-based blocking mechanism and a two-level attention scheme. Nevertheless, the static partitioning of spatial blocks may split spatially adjacent nodes in real traffic networks, thus weakening the model’s ability to capture local spatial dependencies effectively. GSNet \cite{GSNet} leverages the sparsity of the adaptive adjacency matrix to reduce computational complexity by introducing a relation compressor and a feature extractor. However, the compression of spatial relations may result in incomplete modeling of spatial dependencies.\par
Second, in feature-based methods, informative node embeddings can enable accurate predictions without relying on explicit graph convolutions \cite{STID}. However, these embeddings often account for a significant portion of the model parameters, yet few existing approaches consider their regularization. As illustrated in Figure \ref{fig:intro}, we analyze the parameter distribution of GWNet \cite{GWNet}, AGCRN \cite{AGCRN}, STID \cite{STID}, STAEFormer \cite{STAEFormer}, and BigST \cite{BigST} on the LargeST-CA \cite{LargeST} dataset. Among these models, GWNet and AGCRN use node embedding to construct adaptive adjacency matrix, while STID, STAEFormer, and BigST use node embedding as additional features. The results show that node embeddings dominate the number of total parameters in feature-based methods. Therefore, exploring effective regularization strategies for node embeddings is crucial to mitigating overfitting in traffic prediction models. Although certain graph structure regularization techniques have been introduced \cite{RGSL, STC-Dropout}, they primarily focus on the adaptive adjacency matrix and fail to address the fundamental issue of overfitting at the embedding level. Regularization for embeddings, such as Laplacian regularization, directly penalizes the distance between embeddings, but lacks data-driven flexibility and relies on the quality of the prior graph structure. The Stochastic Shared Embedding (SSE) mechanism introduces perturbations by randomly replacing node embeddings during training to achieve regularization. SSE has proven effective in fields such as recommendation systems \cite{SSE-PT}, computer vision \cite{SSE_CV}, and natural language processing \cite{SSE}. However, traffic forecasting is a time series prediction task that is particularly sensitive to signal perturbations. In this context, introducing noise into node embeddings may destabilize the training process and lead to underfitting.\par
To address the aforementioned challenges, we propose Regularized Adaptive Graph Convolution (RAGC) model. For the first issue, we introduce the Efficient Cosine Operator (ECO), a graph convolution module with linear computational complexity. Specifically, in ECO, we construct a semantic adjacency matrix based on the cosine similarity between node embeddings. By decomposing cosine similarity computation into grouped vector multiplications, ECO avoids the explicit construction of an $N\times N$ adjacency matrix, thereby reducing the computational complexity to $O(N)$. For the second issue, we develop a regularized adaptive graph convolution framework that organically integrates SSE with adaptive graph convolution. On one hand, SSE introduces regularized node embeddings to prevent the over-parameterization of the node embeddings that dominate the parameters. On the other hand, by incorporating residual differencing, the adaptive graph convolution filters out the noise introduced by shared embeddings, ensuring that the numerically sensitive spatial-temporal prediction is not adversely affected by such perturbations. In summary, the main contributions of this paper are as follows:\par
\begin{itemize}
\item We propose ECO, a linear complexity graph convolution operator based on cosine similarity of node embeddings, ensuring scalability to large traffic networks without sacrificing spatial expressiveness.
\item We propose a regularized adaptive graph convolution framework that integrates SSE with adaptive graph convolution to achieve regularized graph learning. Simultaneously, a residual difference mechanism is employed to suppress the propagation of stochastic noise, enabling a synergistic interplay between embedding regularization and adaptive graph learning.
\item We validate the proposed RAGC model through extensive experiments on four large-scale traffic flow datasets against 12 widely adopted baseline models. The results demonstrate that RAGC consistently outperforms state-of-the-art methods in terms of prediction accuracy and ranks second and third in training speed and inference speed, respectively.
\end{itemize}\par
The rest of this paper is organized as follows. Section \ref{sec:pre} introduces the problem formulation for traffic forecasting. Section \ref{sec_analysis} discusses the motivation behind our proposed approach. Section \ref{sec:method} provides a detailed description of the proposed methodology. Experimental results on four large-scale traffic forecasting datasets are presented in Section \ref{sec:exp}. Section \ref{sec:related_work} reviews related work, and Section \ref{sec:conclusion} concludes the paper.

\section{Preliminaries}\label{sec:pre}
\subsection{Traffic Network}
We model the traffic network as a graph $G=(V,\mathcal{E},A)$, where $V$ is the set of nodes and $N=|V|$ is the number of nodes, $\mathcal{E}$ is the set of edges, and $A\in \mathbb{R}^{N\times N}$ is the adjacency matrix derived from the traffic network. Typically, a node denotes a sensor located at a specific location within the traffic network. Each sensor records traffic data at certain time intervals. Let $X_t\in \mathbb{R}^{N\times C}$ represent the traffic data of $N$ nodes at time step $t$, where $C$ is the number of features observed by the sensors (e.g., traffic flow, traffic speed, etc.).
\subsection{Problem Definition}
Given the graph $G$ and the historical traffic data from the previous $T$ time steps, we denote the data sequence as ${X}^{{t-T+1}:{t}}=(X^{t-T+1}, \cdots, X^t) \in \mathbb{R}^{N\times T\times C}$, where the notation ${X}^{{t-T+1}:{t}}$ represents the traffic flow observations from time step ${t-T+1}$ through $t$. The goal of traffic forecasting is to find a function $f$ that predicts the traffic data for the next $T'$ steps ${X}^{{t+1}:{t+T'}}\in \mathbb{R}^{N\times T'\times C}$ based on ${X}^{{t-T+1}:{t}}$:
\begin{equation}\label{equ_define}
[X^{t-T+1}, \cdots, X^t;G] \overset{f}{\longrightarrow} [X^{t+1}, \cdots, X^{t+T'};G]
\end{equation}

For ease of reference, Table~\ref{tab:notations} summarizes the notations used throughout this paper.

\begin{table}[ht]
\centering
\caption{Notations and Explanations.}
\label{tab:notations}
\begin{tabular}{cl}
\hline
\textbf{Notations} & \textbf{Explanations} \\
\hline
$V$ & Set of nodes in the road network \\
$\mathcal{E}$ & Set of edges in the road network \\
$A$ & Adjacency matrix of the road network \\
$G = (V, \mathcal{E}, A)$ & Graph representing the road network \\
$N$ & Number of nodes in the road network \\
$C$ & Number of features observed by sensors \\
${X}$ & Historical traffic flow data \\
$\hat{X}$ & Predicted traffic flow data \\
$E, e$ & Trainable embeddings \\
$H, h$ & Hidden states of the model \\
$Z$ & Diffusion steps of diffuse graph convolution \\
$d$ & Dimension of embeddings or features \\
$S$ & Cosine similarity matrix of node embeddings \\
$W, b$ & Trainable projection parameters \\
$\Theta$ & Set of all trainable parameters in RAGC \\
\hline
\end{tabular}
\end{table}

\section{Analysis}\label{sec_analysis}
In this section, we first analyze why applying SSE in isolation fails to regularize traffic flow prediction models, and then we motivate the design of our approach.\par
To learn spatial dependencies in a data‑driven manner, adaptive graph learning introduces a learnable node‐embedding matrix $E\in \mathbb{R}^{N\times d}$, where $N$ is the number of nodes and $d$ is the embedding dimension. Denote the embedding of node $i$ by
\begin{equation}\label{equ_nodei}
e_i \in \mathbb{R}^d, ~ i=1, 2, \cdots, N
\end{equation}

In the SSE \cite{SSE} operation, for each node $i$, a Bernoulli mask $m_i$ is sampled:
\begin{equation}\label{equ_Bernoulli}
m_i \sim Bernoulli(p)
\end{equation}

\noindent where $p\in[0, 1]$ is the probability of sharing. The degree of randomness is explicitly controlled by the hyperparameter $p$, which specifies the probability that each node’s embedding is replaced by that of another randomly selected node. A larger $p$ increases the replacement probability and thus the stochasticity, whereas a smaller $p$ reduces it. Then an index $r(i)$ is chosen uniformly at random from $\{1,2,\cdots ,N\}$ and the perturbed embedding is:
\begin{equation}\label{equ_etilde}
\tilde{e}_i = m_ie_{r(i)}+(1-m_i)e_i
\end{equation}
Since $r(i)$ is uniform, the expectation of $\tilde{e}_i$ is computed as:
\begin{equation}\label{equ_expectation_etilde}\begin{aligned}
\mathbb{E}[\tilde{e}_i] &= p\cdot \frac{1}{N}\sum^N_{j=1} e_j+(1-p)e_i \\
&=p\bar{e} + (1-p)e_i
\end{aligned}\end{equation}

Thus:
\begin{itemize}
    \item If $p=0$, $\tilde{e}_i = e_i$ (no perturbation).
    \item If $p>0$, each embedding retains its own information plus a fraction of the global average $\bar{e}=\frac{1}{N}\sum^N_{j=1}e_j$.
\end{itemize}

By injecting a global average noise, SSE encourages each node to reduce its reliance on its own embedding, making it a useful regularizer in many areas \cite{SSE-PT}. For example, in recommendation systems, graph nodes represent users or items whose embeddings encode rich, semantically diverse signals (preferences, attributes, etc.). Randomly replacing embeddings introduces beneficial semantic mixing, helping the model explore latent relationships without severely harming predictive accuracy. However, traffic flow prediction is fundamentally a spatial-temporal prediction task requiring high continuity and numerical precision. When SSE randomly substitutes node embeddings, it effectively injects spatial-temporal noise into the input signal. Through residual connections and multi‐layer propagation, these perturbations distort the learned time series and can accumulate across layers, ultimately destabilizing training.

Specifically, assume the model consists of $L$ layers, the model of the $l$-th layer is represented as $\mathcal{F}^{(l)}(\cdot)$, and each layer employs residual connections. The layer-wise propagation can be expressed as:
\begin{equation}\label{equ_model_case} 
H^{(l)} = H^{(l-1)} + \mathcal{F}^{(l)}(H^{(l-1)}), \quad l = 1, 2, \ldots, L
\end{equation}
Accordingly, the final output at the $L$-th layer is:
\begin{equation}\label{equ_model_out} 
H^{(L)} = \tilde{e}_i + \sum_{l=1}^{L} \mathcal{F}^{(l)}(H^{(l)}) 
\end{equation}
This formulation indicates that the perturbed embedding $\tilde{e}_i$, which incorporates global average noise, is preserved in the residual path and directly propagated to the output layer. As a result, the injected noise continuously influences the model across all layers, potentially compromising training stability.

To address this issue, we introduce diffusion graph convolution \cite{DGCRN, GWNet} to block the propagation of noise to the deep layers of the model. Assuming that the adjacency matrix used for graph convolution is $A$ and the perturbed node embedding $\tilde{e}$, the diffusion process of graph convolution with 
$Z$ finite steps is:
\begin{equation}\label{equ_analysis_gcn} h_i=\sum_{z=0}^{Z}\sum_{k=1}^{N}A^{(z)}_{ik}\tilde{e}_kW^{(z)}_{g}
\end{equation}

\noindent where $A^{(z)}$ represents the power series of the adjacency matrix, $A^{(0)}=I_{N}$ and $W^{(z)}_{g}$ is a learnable weight matrix. Substituting Equation~(\ref{equ_etilde}) into the above, we obtain:
\begin{equation}
h_i=\sum_{z=0}^{Z}\sum_{k=1}^{N}A^{(z)}_{ik}(m_k \cdot e_{r(k)}+(1-m_k)\cdot e_k)W^{(z)}_{g}
\end{equation}

Taking the expectation of $h_i$, we derive:
\begin{equation}\begin{aligned}
\mathbb{E}[h_i]&= \sum_{z=0}^{Z}\sum_{k=1}^{N}A^{(z)}_{ik}(p\bar{e}+(1-p)\cdot e_k)W^{(z)}_{g} \\
&= p\bar{e}\sum_{z=0}^{Z}\sum_{k=1}^{N}A^{(z)}_{ik}W^{(z)}_{g}+(1-p)\sum_{z=0}^{Z}\sum_{k=1}^{N}A^{(z)}_{ik}e_kW^{(z)}_{g}
\end{aligned}\end{equation}

Since $A$ is row-normalized, i.e., $\sum_{k=1}^{N}A^{(z)}_{ik}=1$, it follows that:
\begin{equation}
\mathbb{E}[h_i]= p\bar{e}\sum_{z=0}^{Z}W^{(z)}_{g}+(1-p)\sum_{z=0}^{Z}\sum_{k=1}^{N}A^{(z)}_{ik}e_kW^{(z)}_{g}
\end{equation}

Next, consider the expectation of the difference between the perturbed embedding and the graph convolution output (assuming the input and output dimensions of $W^{(z)}_{g}$ are the same):
\begin{equation}\begin{aligned}\label{equ_analysis_diff}
\mathbb{E}[\tilde{e}_i-h_i]&= p\bar{e} + (1-p)e_i\\
&~~~~-p\bar{e}\sum_{z=0}^{Z}W^{(z)}_{g}-(1-p)\sum_{z=0}^{Z}\sum_{k=1}^{N}A^{(z)}_{ik}e_kW^{(z)}_{g} \\
&=p\bar{e}(I-\sum_{z=0}^{Z}W_{g}^{(z)})\\
&~~~~+(1-p)(e_i-\sum_{z=0}^{Z}\sum_{k=1}^{N}A^{(z)}_{ik}e_kW^{(z)}_{g})
\end{aligned}\end{equation}

This result shows that the global average noise term $\bar{e}$ can be effectively mitigated by the learnable weight matrix $W_{g}$. Specifically, the graph convolution operation can adaptively determine whether to suppress the noise in a data-driven manner. If the model opts to eliminate the noise, the sum of the weights of each diffusion step, $\tilde{W_{g}}=\sum_{z=0}^{Z} W_{g}^{(z)}$, should approximate the identity matrix $I$, and vice versa. Through this subtraction mechanism, graph convolution not only prevents the propagation of noise to deeper layers, but also preserves the aggregated information from each node and its neighbors. As a result, it achieves regularization while maintaining the training stability necessary for time series forecasting.

\section{Methodology}\label{sec:method}
\subsection{Overview}
In this section, we introduce the proposed RAGC framework in detail. As shown in Figure~\ref{fig:model}, RAGC primarily consists of three components, namely the embedding layer, the adaptive graph convolution encoder, and the regression layer. The embedding layer transforms the historical traffic sequences into high-dimensional representations and incorporates time embeddings and node embeddings as additional features. Subsequently, the adaptive graph convolution encoder processes these features to capture spatial-temporal dependencies. Finally, the regression layer projects the encoded features to the predicted traffic sequences. The remainder of this section provides a detailed description of each component.

\subsection{Embedding Layer}
\subsubsection{Spatial-Temporal Embedding}

Following prior studies \cite{STID, STAEFormer, PatchSTG}, we first transform the input historical traffic data ${X}^{{t-T+1}:{t}}\in \mathbb{R}^{N\times (T\times C)}$ into a high-dimensional input embedding ${E}_{in}\in \mathbb{R}^{N\times d_{in}}$ via a fully connected layer:
\begin{equation}\label{equ_data_emb}
{E}_{in} = W{X}^{{t-T+1}:{t}}+b
\end{equation}

\noindent where $W\in \mathbb{R}^{(T\times C)\times d_{in}}$, $b\in \mathbb{R}^{d_{in}}$ are learnable parameters of the fully connected layer. To capture temporal heterogeneity, we introduce two learnable embedding dictionaries, namely ${D}_{tid} \in \mathbb{R}^{T_d \times d_{tid}}$ for time-of-day and ${D}_{diw} \in \mathbb{R}^{T_w \times d_{diw}}$ for day-of-week representations, where $T_d$ and $T_w$ denote the number of time steps per day and per week, respectively. We use the timestamp of the last input step as the query to extract the corresponding time-of-day embedding ${E}_{tid}$ and day-of-week embedding ${E}_{diw}$. For adaptive graph learning, we further define learnable node embeddings ${E}_{node} \in \mathbb{R}^{N \times d_{node}}$, where $d_{node}$ is the dimension of node embeddings.

\begin{figure}[ht]
  \centering
  \includegraphics[width=0.98\linewidth]{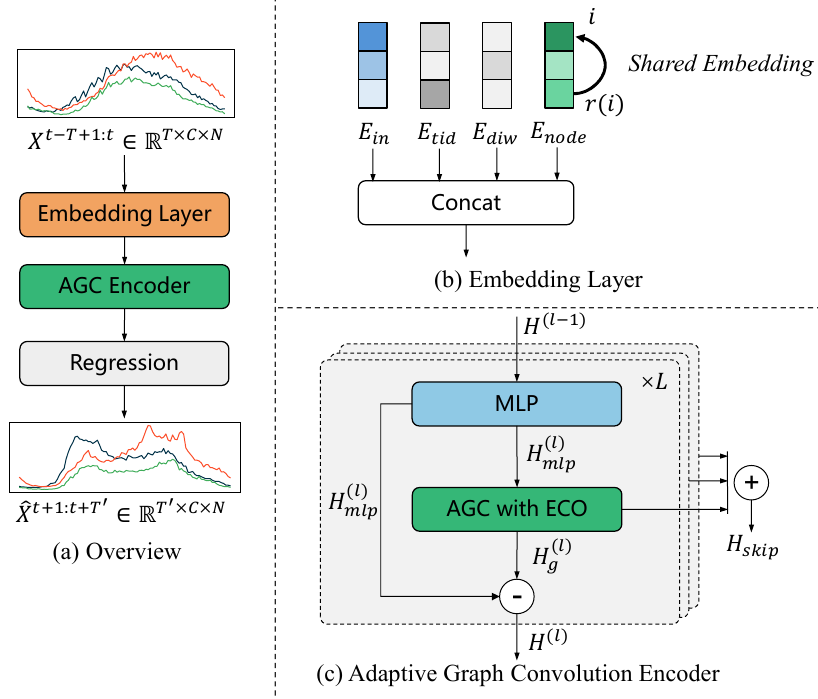}
  \caption{Detailed framework of RAGC.}
  \label{fig:model}
\end{figure}

\subsubsection{SSE for Node Embedding Regularization}
To improve the generalization of node embeddings and mitigate overfitting, we adopt SSE \cite{SSE}, which randomly replaces the embeddings between nodes during training. As mentioned in Section \ref{sec_analysis}, this procedure yields a regularized embedding matrix $\tilde{E}_{node} \in \mathbb{R}^{N \times d_{node}}$, which is then fed into the subsequent graph convolution layers. As illustrated in Figure~\ref{fig:model}, when the Bernoulli sampling in Equation~\ref{equ_Bernoulli} yields $m_i=1$ for node $i$, the embedding of node $i$ is replaced by that of another randomly selected node $r(i)$. By sharing embeddings between global nodes, SSE introduces greater stochasticity and breaks the constraints imposed by local structural dependencies. This results in improved regularization and better generalization performance of the node embedding.

\subsubsection{Output of Embedding Layer}
Finally, we concatenate the above embeddings to obtain the regularized spatial-temporal embedding as input to the adaptive graph convolution encoder:
\begin{equation}\label{equ_concat_emb}
    {H}^{(0)} = {E}_{in} \parallel {E}_{tid} \parallel {E}_{diw} \parallel \tilde{E}_{node}
\end{equation}

\noindent where ${H}^{(0)}\in \mathbb{R}^{N\times d_{0}}$ and $d_{0}=d_{in}+d_{tid}+d_{diw}+d_{node}$.

\subsection{Adaptive Graph Convolution Encoder}

\subsubsection{Multi-Layer Perceptron} with Residual Connections
At each layer, we first apply a Multi-Layer Perceptron (MLP) with residual connection to the spatial-temporal embeddings to enable feature interaction and nonlinear transformation, thereby capturing complex spatial-temporal dependencies:
\begin{equation}\label{equ_mlp}
{H}^{(l)}_{mlp}=\mathrm{FC}_{2}(\mathrm{ReLU} (\mathrm{FC}_{1}({H}^{(l-1)}))) + {H}^{(l-1)}
\end{equation}
\noindent where $\mathrm{FC}(\cdot)$ denotes a fully connected layer, ${H}^{(l)}_{mlp}\in\mathbb{R}^{N\times d_{0}}$ is the output.

\subsubsection{Residual Difference Mechanism}
As discussed in Section~\ref{sec_analysis}, residual connections may propagate the noise introduced by SSE into deeper layers, potentially destabilizing training. To address this, we introduce an adaptive graph convolution to aggregate spatial information:
\begin{equation}\label{equ_gcn}
{H}^{(l)}_{g}=\sum_{z=0}^{Z}A^{(z)}_{adp}{H}^{(l)}_{mlp}W^{(z)}_{g}
\end{equation}

\noindent where $Z$ is the number of diffusion steps of graph convolution, $A_{adp}$ is the adaptive adjacency matrix and $W^{(z)}_g\in\mathbb{R}^{d_{0}\times d_{0}}$ is a learnable weight matrix. To suppress the global noise introduced by SSE, we compute the residual difference between the original and aggregated representations:
\begin{equation}\label{equ_res_diff}
{H}^{(l)}={H}^{(l)}_{mlp}-{H}^{(l)}_{g}
\end{equation}

\noindent where ${H}^{(l)}\in\mathbb{R}^{N\times d_{0}}$ serves as the output for the $l$-th layer. As discussed in Section~\ref{sec_analysis}, the residual difference mechanism enables the learnable weight matrices $W^{(z)}_g$ in Equation~\ref{equ_gcn} to adaptively determine whether to suppress noise in a data-driven manner, thereby reducing the noise present in ${H}^{(l)}$. From a signal decomposition perspective, the smoothed representation ${H}^{(l)}_g$ reflects the global trend among node neighbors, while the residual ${H}^{(l)}$ captures node-level fluctuations. Together, they form the traffic flow pattern of each node. To retain both components, we introduce a skip connection by defining ${H}_{skip} = \sum_{l=1}^{L}{H}^{(l)}_g \in \mathbb{R}^{N\times d_{0}}$, which is later incorporated into the final output instead of being discarded.

\begin{figure}[ht]
  \centering
  \includegraphics[width=0.98\linewidth]{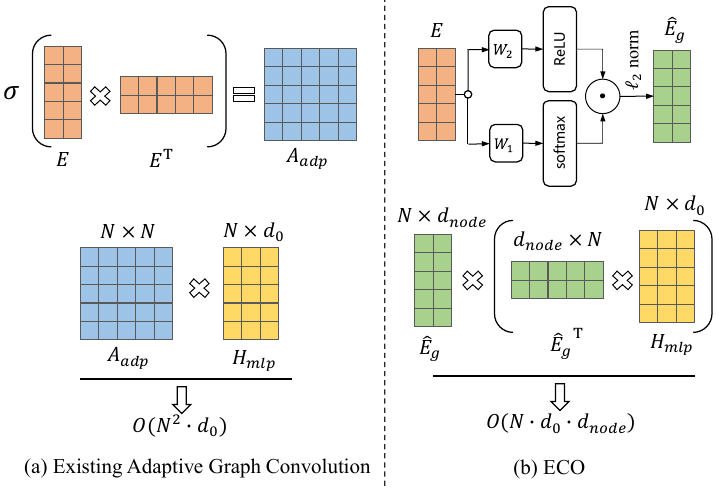}
  \caption{Comparison of (a) Existing Adaptive Graph Convolution and (b) our proposed ECO.}
  \label{fig:ECO}
\end{figure}

\subsubsection{Efficient Cosine Operator}
In prior adaptive graph learning approaches \cite{GWNet, AGCRN}, the adaptive adjacency matrix $A_{adp}$ is typically computed via the Gram matrix of node embeddings, followed by row-wise 
 normalization with a softmax function:
\begin{equation}\label{equ_adp_gwnet}
A_{adp}=\mathrm{softmax}(\mathrm{ReLU}(E_{node}E_{node}^{\mathrm{T}}))
\end{equation}

\noindent where ReLU is used to suppress weak connections. This formulation avoids learning a full $N \times N$ adjacency matrix, relying instead on $NC$ node embedding parameters. However, as illustrated in Figure \ref{fig:ECO}(a), the graph convolution operation ${A}_{adp}{H}_{mlp}^{(l)}$ still incurs $O(N^2)$ complexity, limiting scalability in large-scale road networks. The main bottleneck arises from the non-decomposability of the softmax and ReLU functions, which prevents reordering of matrix multiplications for computational efficiency. If the computation of the normalized Gram matrix can be decomposed, the node embedding $E_{node}$ can be directly multiplied with the node features $H_{mlp}^{(l)}$ by leveraging the associative property of matrix multiplication. This enables the graph convolution to be reformulated in a manner that reduces the computational complexity from $O(N^2)$ to $O(N)$, as illustrated in Figure \ref{fig:ECO}(b). \par

To address this, BigST \cite{BigST} applies random feature mapping to approximate softmax and and make it multiplicatively decomposable, but at the cost of additional approximation noise that can impair prediction performance. In contrast, we argue that the essential purpose of $A_{adp}$ is to measure node similarity, which can be effectively captured via cosine similarity—without nonlinearities.\par

We introduce ECO, a linear-complexity graph convolution operator that leverages cosine similarity. First, we apply a gating mechanism to transform node embeddings, ensuring non-negativity and enabling selective connectivity:
\begin{equation}\label{equ_embtrans}
E_{g} = \mathrm{softmax}(E_{node}W_{1}) \odot \mathrm{ReLU}(E_{node} W_{2})
\end{equation}

\noindent where $W_1, W_2 \in \mathbb{R}^{d_{node} \times d_{node}}$ are learnable parameters, and $\odot$ denotes element-wise multiplication. The intuition behind this gating operation is that connections between graph nodes should remain sufficiently sparse. To this end, we apply a softmax activation to emphasize the most relevant neighbors, and a ReLU activation to remove weak connections with negative values, ensuring the adjacency matrix remains non-negative. These connection relationships are learned in a data-driven manner through the trainable parameters $W_1$ and $W_2$.\par

Next, we compute the similarity matrix $S \in \mathbb{R}^{N \times N}$ by applying cosine similarity to the $\ell_2$-normalized node embeddings:
\begin{equation}\label{equ_cossim} 
S = \hat{E}_g \hat{E}_g^\mathrm{T}, \quad \hat{E}_g = \frac{E_g}{\|E_{g}\|_2} 
\end{equation}

\noindent The adaptive adjacency matrix is then obtained via row-wise normalization:
\begin{equation}\label{equ_gcn_adp}
\begin{aligned}
A_{adp} &= D^{-1}S \\
D &= diag(S1_N)
\end{aligned}
\end{equation}

\noindent where $1_N$ is an $N$-dimensional vector of ones, $diag(\cdot)$ is the diagonal matrix conversion function and $D$ is the degree matrix. In our approach, $S$ naturally includes self-similarity, because the cosine similarity of a vector with itself is always 1. We intentionally preserve this self-similarity to retain each node’s original feature information. Substituting Equation (\ref{equ_cossim}) into the graph convolution yields:
\begin{equation}\label{equ_gcn_flatten}
{A}_{adp}{H}_{mlp}^{(l)}=D^{-1}\hat{E_{g}}\hat{E_{g}}^{\mathrm{T}}{H}_{mlp}^{(l)}
\end{equation}
Following the matrix associativity trick, we can rewrite the operation as:
\begin{equation}\label{equ_linearized_gcn_asso}
\begin{aligned}
{A}_{adp}{H}_{mlp}^{(l)} &= D^{-1}(\hat{E_{g}}(\hat{E_{g}}^{\mathrm{T}}{H}_{mlp}^{(l)}))\\
D &= diag(\hat{E_{g}}(\hat{E_{g}}^{\mathrm{T}}1_N))
\end{aligned}
\end{equation}

These transformations eliminate the need to explicitly construct the $N \times N$ adjacency matrix $A_{adp}$ through pairwise similarity calculations, which would otherwise incur a computational complexity of $O(N^2 \cdot d_{0})$ in the feature aggregation operation. Instead, we leverage the vector $\hat{E}_{g}$ directly in the computation, reducing the complexity to $O(N \cdot d_{0} \cdot d_{node})$. As a result, the overall computational cost of graph convolution is effectively reduced from $O(N^2)$ to $O(N)$, as illustrated in Figure~\ref{fig:ECO}(b). Unlike the method in \cite{BigST}, which relies on random feature map approximations, our approach derives $\hat{E}_{g}$ directly from cosine similarity, thereby avoiding the additional noise introduced by such approximations.

\subsection{Regression Layer}

After passing through the $L$-layer adaptive graph convolution encoder, we obtain the hidden representation $H^{(L)}$, which captures node-level fluctuations, and $H_{skip}$, which encodes the global trend within each node’s neighborhood. These two representations are then separately processed by fully connected layers for regression. The final prediction is obtained by summing the outputs of the two branches:
\begin{equation}\label{equ_reg}
\hat{X}^{t+1:t+T'} = \mathrm{FC}_{node}(H^{(L)}) + \mathrm{FC}_{global}(H_{skip})
\end{equation}

\subsection{Loss Function}
The Mean Absolute Error (MAE) is selected as the loss function:
\begin{equation}\label{equ_mae}
\mathcal{L}(\hat{X}^{t+1:t+T'};\Theta)=\frac{1}{NT'C}\sum^{N}_{i=1}\sum^{T'}_{j=1}\sum^{C}_{c=1}|\hat{X}^{t+j}_{ic}-X^{t+j}_{ic}|
\end{equation}
where $\Theta$ represents all trainable parameters of RAGC, $\hat{X}^{t+j}_{ic}$ is the prediction of the model, and $X^{t+j}_{ic}$ is the ground truth.\par
The forward propagation process of RAGC is summarized in Algorithm~\ref{alg:ragl}. Lines 1–4 describe the operations in the embedding layer. Line 5 presents the gating mechanism employed in the ECO module. Lines 6–12 detail the operations of the adaptive graph convolution encoder. Line 13 illustrates the operations performed by the output layer.

\begin{algorithm}[h]
\caption{The forward propagation process of RAGC}
\small
\label{alg:ragl}
\KwIn{Historical data ${X}^{{t-T+1}:{t}} \in \mathbb{R}^{N\times T\times C}$}
\KwOut{Predictions $\hat{X}^{t+1:t+T'}\in \mathbb{R}^{N\times T'\times C}$ based on ${X}^{{t-T+1}:{t}}$}

${E}_{in} \leftarrow {E}_{in} = W{X}^{{t-T+1}:{t}}+b$ \tcp*[r]{Time-series embedding}

${E}_{tid} \leftarrow {D}_{tid}[\text{time-of-day timestamp}]$\;
${E}_{diw} \leftarrow {D}_{diw}[\text{day-of-week timestamp}]$\;

$\tilde{E}_{node} \leftarrow \text{SSE}({E}_{node})$ \tcp*[r]{Stochastic shared node embedding}

$\hat{E}_g \leftarrow \mathrm{Normalize}(\mathrm{softmax}(E_{node}W_{1}) \odot \mathrm{ReLU}(E_{node}W_{2}))$ \tcp*[r]{Gating mechanism}

${H}^{(0)} \leftarrow \mathrm{Concat}({E}_{in}, {E}_{tid}, {E}_{diw}, \tilde{E}_{node})$\;
${H}_{skip} \leftarrow 0$\;

\For{$l=1$ \KwTo $L$}{
    ${H}^{(l)}_{mlp} \leftarrow \mathrm{MLP}({H}^{(l-1)})$\;
    ${H}^{(l)}_{g} \leftarrow \mathrm{ECO}({H}^{(l)}_{mlp}, \hat{E}_g)$\;
    ${H}^{(l)} \leftarrow {H}^{(l)}_{mlp} - {H}^{(l)}_{g}$\ \tcp*[r]{Residual Difference Mechanism}
    ${H}_{skip} \leftarrow {H}_{skip} + {H}^{(l)}_{g}$\;
}

$\hat{X}^{t+1:t+T'} \leftarrow \mathrm{FC}_{node}(H^{(L)}) + \mathrm{FC}_{global}(H_{skip})$

\Return $\hat{X}^{t+1:t+T'}$
\end{algorithm}

\subsection{Complexity Analysis}
This section analyzes the computational complexity of the RAGC framework. For simplicity, we assume that $d_{in}$, $d_{node}$, and $d_{0}$ are all $O(d)$ in the following analysis. In the data embedding layer, the computational complexity of converting the historical input data into input embeddings is $O(N\cdot T\cdot d)$. In the adaptive graph convolution encoder, the MLP encoding has a complexity of $O(N\cdot d^2)$. The computational complexity of computing $\hat{E_g}$ and performing linearized graph convolution in the ECO module is $O(N\cdot d^{2})$. In the output layer, converting the hidden state into the predicted sequence requires a complexity of $O(N\cdot T'\cdot d)$. For RAGC with $L$ stacked layers, the total computational complexity is $O(L\cdot N \cdot d \cdot (T+d+T'))$. Since $T$, $T'$, $L$, and $d$ are fixed hyper-parameters, the overall computational complexity is linear with respect to the number of road network nodes $N$.

\section{Experiments}\label{sec:exp}
\subsection{Experimental Setup}

\subsubsection{Datasets}
We conducted experiments on the large-scale traffic flow prediction dataset LargeST \cite{LargeST}, which includes 8,600 sensors across California and consists of four sub-datasets, namely SD, GBA, GLA, and CA. Following prior works \cite{RPMixer, PatchSTG}, we split each dataset into training, validation, and test sets with a ratio of 6:2:2. Both the historical input length and the prediction horizon are set to 12 time steps (i.e., $T = T' = 12$). Detailed statistics of all sub-datasets are provided in Table~\ref{tab:table_dataset}.\par
We follow the procedure described in \cite{DCRNN, LargeST} to compute the pairwise road network distances between sensors and to construct a spatial distance-based adjacency matrix $A_{geo}$. Specifically, a thresholded Gaussian kernel \cite{gaussian} is applied, where $A_{ij} = \exp(-\frac{r_{ij}^2}{\sigma^2})$ if $\exp(-\frac{r_{ij}^2}{\sigma^2}) \ge \epsilon$, and $A_{ij} = 0$ otherwise. Here, $A_{ij}$ represents the edge weight between node $i$ and node $j$, and $r_{ij}$ denotes the road network distance between them. The parameter $\sigma$ is the standard deviation of the distance distribution, and $\epsilon$ is the threshold for sparsification. 

\begin{table}[!h]\centering
    \caption{\normalsize Dataset Description (1M = 1,000,000).}
    \label{tab:table_dataset}
    \resizebox{0.47\textwidth}{!}{
    \begin{tabular}{*{5}{c}}
        \toprule
       Datasets & Nodes & Records & Sample Rate & Timespan \\
        \midrule
        SD & 716 & 25M & 15min & 01/01/2019-12/31/2019 \\
        GBA & 2,352 & 82M & 15min & 01/01/2019-12/31/2019 \\
        GLA & 3,834 & 134M & 15min & 01/01/2019-12/31/2019 \\
        CA & 8,600 & 301M & 15min & 01/01/2019-12/31/2019 \\
        \bottomrule
    \end{tabular}
    }
\end{table}

\subsubsection{Baselines}
We compare RAGC against 12 widely adopted baseline models, categorized into three groups: (i) STGNN-based methods: GWNet \cite{GWNet}, AGCRN \cite{AGCRN}, STGODE \cite{STGODE}, DSTAGNN \cite{DSTAGNN}, D2STGNN \cite{D2STGNN}, DGCRN \cite{DGCRN}, BigST \cite{BigST}, and GSNet \cite{GSNet}. (ii) MLP-based methods: STID \cite{STID} and RPMixer \cite{RPMixer}. (iii) Attention-based methods: STWave \cite{STWave} and PatchSTG \cite{PatchSTG}. Among these, GWNet, AGCRN, and D2STGNN construct adaptive adjacency matrices using node embeddings, while STID and PatchSTG incorporate node embeddings as additional features. Notably, BigST, GSNet, STWave, and PatchSTG explicitly address model scalability by optimizing computational complexity.

\subsubsection{Metrics}
Three metrics are used in the experiments: (1) Mean Absolute Error (MAE), (2) Root Mean Squared Error (RMSE), (3) Mean Absolute Percentage Error (MAPE). Mean Absolute Error (MAE) measures the average magnitude of the absolute differences between predicted values and the ground truth, providing a straightforward evaluation of prediction accuracy. Root Mean Squared Error (RMSE), in contrast, assigns greater weight to larger errors by squaring the differences before averaging, making it particularly sensitive to significant deviations in predictions. Finally, Mean Absolute Percentage Error (MAPE) expresses the error as a percentage of the ground truth, facilitating relative comparisons across datasets or models with varying scales. Collectively, these metrics offer a comprehensive evaluation of the accuracy and robustness of the model’s predictions.

\subsubsection{Implementation Details}
We implement RAGC using PyTorch 1.10.0 with Python 3.8.10. All experiments are conducted on a server equipped with an Intel(R) Xeon(R) Platinum 8255C CPU @ 2.50GHz, 43 GB of RAM, and a single NVIDIA GeForce RTX 3090 GPU with 24 GB of memory. We set the $d_{in}$, $d_{tid}$ and $d_{diw}$ as $32$. The node embedding dimension $d_{node}$ is searched over the set $\{32, 48, 64, 80\}$. The diffusion step $Z$ is set as 1 to avoid over-smoothing. The number of encoder layers $L$ is searched over $\{2, 4, 6, 8\}$, and the replacement probability of the SSE mechanism $p$ is selected from $\{0.05, 0.1, 0.2, 0.3\}$. We train the model for 200 epochs using the Adam optimizer with an initial learning rate of 0.002, which decays by a factor of 0.5 every 40 epochs. The batch size is set to 64 by default. In cases of GPU memory overflow, the batch size is progressively reduced until the model can run without memory constraints.

\begin{table*}[h!]
\caption{Comparison of RAGC and baselines on four large-scale traffic forecasting datasets.}
\label{tab:benchmark}
\begin{threeparttable}
\centering
\begin{tabular}{c|l|ccc|ccc|ccc|ccc}
\toprule[1.5pt]
\multirow{2}{*}{Datasets} & \multirow{2}{*}{Methods} & \multicolumn{3}{c|}{Horizon 3} & \multicolumn{3}{c|}{Horizon 6} & \multicolumn{3}{c|}{Horizon 12} & \multicolumn{3}{c}{Average} \\ 
\cline{3-14}
 &  & MAE & RMSE & MAPE & MAE & RMSE & MAPE & MAE & RMSE & MAPE & MAE & RMSE & MAPE \\ \hline \hline
\multirow{12}{*}{SD} 
 & GWNet & 15.24 & 25.13 & 9.86\% & 17.74 & 29.51 & 11.70\% & 21.56 & 36.82 & 15.13\% & 17.74 & 29.62 & 11.88\% \\
 & AGCRN & 15.71 & 27.85 & 11.48\% & 18.06 & 31.51 & 13.06\% & 21.86 & 39.44 & 16.52\% & 18.09 & 32.01 & 13.28\% \\ 
 & STGODE & 16.75 & 28.04 & 11.00\% & 19.71 & 33.56 & 13.16\% & 23.67 & 42.12 & 16.58\% & 19.55 & 33.57 & 13.22\% \\
 & DSTAGNN & 18.13 & 28.96 & 11.38\% & 21.71 & 34.44 & 13.93\% & 27.51 & 43.95 & 19.34\% & 21.82 & 34.68 & 14.40\% \\
 & D$^2$STGNN & 14.92 & 24.95 & 9.56\% & 17.52 & 29.24 & 11.36\% & 22.62 & 37.14 & 14.86\% & 17.85 & 29.51 & 11.54\% \\
 & DGCRN & 15.34 & 25.35 & 10.01\% & 18.05 & 30.06 & 11.90\% & 22.06 & 37.51 & 15.27\% & 18.02 & 30.09 & 12.07\% \\
 & BigST & 16.42 & 26.99 & 10.86\% & 18.88 & 31.60 & 13.24\% & 23.00 & 38.59 & 15.92\% & 18.80 & 31.73 & 12.91\% \\
 & GSNet & 17.33 & 27.62 & 10.88\% & 21.69 & 34.01 & 13.75\% & 30.95 & 45.82 & 20.20\% & 22.38 & 34.43 & 14.23\% \\
 & STID & 15.15 & 25.29 & 9.82\% & 17.95 & 30.39 & 11.93\% & 21.82 & 38.63 & 15.09\% & 17.86 & 31.00 & 11.94\% \\
 & RPMixer & 18.54 & 30.33 & 11.81\% & 24.55 & 40.04 & 16.51\% & 35.90 & 58.31 & 27.67\% & 25.25 & 42.56 & 17.64\% \\
 & STWave & 15.80 & 25.89 & 10.34\% & 18.18 & 30.03 & 11.96\% & 21.98 & 36.99 & 15.30\% & 18.22 & 30.12 & 12.20\% \\
 & PatchSTG & \underline{14.53} & \underline{24.34} & \underline{9.22\%} & \underline{16.86} & \underline{28.63} & \underline{11.11\%} & \underline{20.66} & \underline{36.27} & \underline{14.72\%} & \underline{16.90} & \underline{29.27} & \underline{11.23\%} \\
 \cline{2-14}
 & RAGC & \textbf{13.87} & \textbf{23.42} & \textbf{9.01\%} & \textbf{16.09} & \textbf{27.35} & \textbf{10.63\%} & \textbf{19.90} & \textbf{33.94} & \textbf{13.35\%} & \textbf{16.16} & \textbf{27.40} & \textbf{10.62\%} \\
\hline  \hline
\multirow{12}{*}{GBA} 
 & GWNet & 17.85 & 29.12 & 13.92\% & 21.11 & 33.69 & 17.79\% & 25.58 & 40.19 & 23.48\% & 20.91 & 33.41 & 17.66\% \\
 & AGCRN & 18.31 & 30.24 & 14.27\% & 21.27 & 34.72 & 16.89\% & 24.85 & 40.18 & 20.80\% & 21.01 & 34.25 & 16.90\% \\
 & STGODE & 18.84 & 30.51 & 15.34\% & 22.04 & 35.61 & 18.42\% & 26.22 & 42.90 & 22.83\% & 21.79 & 35.37 & 18.26\% \\
 & DSTAGNN & 19.73 & 31.39 & 15.42\% & 24.21 & 37.70 & 20.99\% & 30.12 & 46.40 & 28.16\% & 23.82 & 37.29 & 20.16\% \\
 & D$^2$STGNN & 17.54 & 28.94 & \underline{12.12\%} & 20.92 & 33.92 & 14.89\% & 25.48 & 40.99 & 19.38\% & 20.71 & 33.65 & 15.04\% \\
 & DGCRN & 18.02 & 29.49 & 14.13\% & 21.08 & 34.03 & 16.94\% & 25.25 & 40.63 & 21.15\% & 20.91 & 33.83 & 16.88\% \\
 & BigST & 18.70 & 30.27 & 15.55\% & 22.21 & 35.33 & 18.54\% & 26.98 & 42.73 & 23.68\% & 21.95 & 35.54 & 18.50\% \\
 & GSNet & 18.75 & 30.30 & 14.49\% & 22.59 & 36.07 & 18.18\% & 27.57 & 43.41 & 22.32\% & 22.27 & 35.56 & 17.68\% \\
 & STID & 17.36 & 29.39 & 13.28\% & 20.45 & 34.51 & 16.03\% & 24.38 & 41.33 & 19.90\% & 20.22 & 34.61 & 15.91\% \\
 & RPMixer & 20.31 & 33.34 & 15.64\% & 26.95 & 44.02 & 22.75\% & 39.66 & 66.44 & 37.35\% & 27.77 & 47.72 & 23.87\% \\
 & STWave & 17.95 & 29.42 & 13.01\% & 20.99 & 34.01 & 15.62\% & 24.96 & 40.31 & 20.08\% & 20.81 & 33.77 & 15.76\% \\
 & PatchSTG & \underline{16.81} & \underline{28.71} & 12.25\% & \underline{19.68} & \underline{33.09} & \underline{14.51\%} & \underline{23.49} & \underline{39.23} & \underline{18.93\%} & \underline{19.50} & \underline{33.16} & \underline{14.64\%} \\
 \cline{2-14}
 & RAGC & \textbf{15.71} & \textbf{27.58} & \textbf{10.29\%} & \textbf{18.40} & \textbf{31.89} & \textbf{12.23\%} & \textbf{22.48} & \textbf{38.39} & \textbf{15.92\%} & \textbf{18.33} & \textbf{31.65} & \textbf{12.18\%} \\
 \hline \hline
 \multirow{10}{*}{GLA} 
 & GWNet & 17.28 & 27.68 & 10.18\% & 21.31 & 33.70 & 13.02\% & 26.99 & 42.51 & 17.64\% & 21.20 & 33.58 & 13.18\% \\
 & AGCRN & 17.27 & 29.70 & 10.78\% & 20.38 & 34.82 & 12.70\% & 24.59 & 42.59 & 16.03\% & 20.25 & 34.84 & 12.87\% \\
 & STGODE & 18.10 & 30.02 & 11.18\% & 21.71 & 36.46 & 13.64\% & 26.45 & 45.09 & 17.60\% & 21.49 & 36.14 & 13.72\% \\
 & DSTAGNN & 19.49 & 31.08 & 11.50\% & 24.27 & 38.43 & 15.24\% & 30.92 & 48.52 & 20.45\% & 24.13 & 38.15 & 15.07\% \\
 & BigST & 18.38 & 29.40 & 11.68\% & 22.22 & 35.53 & 14.48\% & 27.98 & 44.74 & 19.65\% & 22.08 & 36.00 & 14.57\% \\
 & GSNet & 18.35 & 29.17 & 11.06\% & 22.58 & 35.71 & 14.00\% & 28.13 & 43.97 & 19.24\% & 22.30 & 35.22 & 14.21\% \\
 & STID & 16.54 & 27.73 & 10.00\% & 19.98 & 34.23 & 12.38\% & 24.29 & 42.50 & 16.02\% & 19.76 & 34.56 & 12.41\% \\
 & RPMixer & 19.94 & 32.54 & 11.53\% & 27.10 & 44.87 & 16.58\% & 40.13 & 69.11 & 27.93\% & 27.87 & 48.96 & 17.66\% \\
 & STWave & 17.48 & 28.05 & 10.06\% & 21.08 & 33.58 & 12.56\% & 25.82 & 41.28 & 16.51\% & 20.96 & 33.48 & 12.70\% \\
 & PatchSTG & \underline{15.84} & \underline{26.34} & \underline{9.27\%} & \underline{19.06} & \underline{31.85} & \underline{11.30\%} & \underline{23.32} & \underline{39.64} & \underline{14.60\%} & \underline{18.96} & \underline{32.33} & \underline{11.44\%} \\
 \cline{2-14}
 & RAGC& \textbf{15.06} & \textbf{25.66} & \textbf{8.39\%} & \textbf{17.84} & \textbf{30.24} & \textbf{10.09\%} & \textbf{21.72} & \textbf{36.73} & \textbf{12.98\%} & \textbf{17.75} & \textbf{30.11} & \textbf{10.20\%} \\
 \hline \hline
 \multirow{8}{*}{CA}
 & GWNet & 17.14 & 27.81 & 12.62\% & 21.68 & 34.16 & 17.14\% & 28.58 & 44.13 & 24.24\% & 21.72 & 34.20 & 17.40\% \\
 & STGODE & 17.57 & 29.91 & 13.91\% & 20.98 & 36.62 & 16.88\% & 25.46 & 45.99 & 21.00\% & 20.77 & 36.60 & 16.80\% \\
 & BigST & 17.15 & 27.92 & 13.03\% & 20.44 & 33.16 & 15.87\% & 25.49 & 41.09 & 20.97\% & 20.32 & 33.45 & 15.91\% \\
 & GSNet & 17.31 & 27.69 & 12.71\% & 20.79 & 32.99 & 15.78\% & 25.41 & 40.01 & 19.68\% & 20.52 & 32.58 & 15.47\% \\
 & STID & 15.51 & 26.23 & 11.26\% & 18.53 & 31.56 & 13.82\% & 22.63 & 39.37 & 17.59\% & 18.41 & 32.00 & 13.82\% \\
 & RPMixer & 18.18 & 30.49 & 12.86\% & 24.33 & 41.38 & 18.34\% & 35.74 & 62.12 & 30.38\% & 25.07 & 44.75 & 19.47\% \\
 & STWave & 16.77 & 26.98 & 12.20\% & 18.97 & 30.69 & 14.40\% & 25.36 & 38.77 & 19.01\% & 19.69 & 31.58 & 14.58\% \\
 & PatchSTG & \underline{14.69} & \underline{24.82} & \underline{10.51\%} & \underline{17.41} & \underline{29.43} & \underline{12.83\%} & \underline{21.20} & \underline{36.13} & \underline{16.00\%} & \underline{17.35} & \underline{29.79} & \underline{12.79\%} \\
 \cline{2-14}
 & RAGC& \textbf{13.98} & \textbf{24.22} & \textbf{9.12\%} & \textbf{16.42} & \textbf{28.34} & \textbf{10.85\%} & \textbf{20.03} & \textbf{34.24} & \textbf{13.83\%} & \textbf{16.40} & \textbf{28.23} & \textbf{10.96\%} \\
\bottomrule[1.5pt]
\end{tabular}
 \begin{tablenotes}
        \footnotesize
        \item  The best result is in bold, and the second best result is underlined.
      \end{tablenotes}
  \end{threeparttable}
\end{table*}

\subsection{Performance Comparison}
Table~\ref{tab:benchmark} presents the comparison results across four large-scale traffic prediction datasets. Some baseline models are missing on the GLA and CA datasets due to out-of-memory issues, even when the batch size is reduced to 4. We evaluate the prediction performance at horizons 3, 6, and 12, as well as the average performance over all time steps. From the results, we draw the following key observations: (1) RAGC consistently outperforms all state-of-the-art baseline models across all datasets and evaluation metrics, demonstrating its effectiveness. The combination of SSE and adaptive graph convolution enables both regularization and noise suppression of node embeddings, leading to more accurate traffic flow predictions. (2) Among STGNN-based methods, GWNet, AGCRN, D2STGNN, and DGCRN are strong baselines. However, several of these models fail to scale to large road networks due to the quadratic computational complexity of graph operations, resulting in infeasible memory requirements. (3) Among MLP-based models, STID achieves competitive results by incorporating node embeddings as input features. However, it does not explicitly regularize these embeddings, limiting its potential for further accuracy improvements. (4) BigST, GSNet, STWave, and PatchSTG consider model scalability and can run on all large-scale datasets without memory overflow. Nevertheless, their reliance on approximations, spatial compression, low-rank factorization, or block partitioning often leads to a loss of spatial information, weakening the ability to capture local spatial dependencies. In contrast, the proposed ECO framework preserves complete spatial information through cosine similarity, thereby delivering superior prediction performance in large-scale scenarios.

\subsection{Ablation Study}
To verify the effectiveness of each component of RAGC, we compare RAGC with seven different variants:
\begin{itemize}
    \item “w/o SSE”: This variant removes the SSE used for node embedding regularization.
    \item “w/o RDM”: This variant removes the Residual Difference Mechanism (RDM), and the output of the adaptive graph convolution will be directly used as the input of the next layer encoder.
    \item “w/o AGC”: This variant replaces the Adaptive Graph Convolution (AGC) with MLP.
    \item “w/ Dropout”: This variant uses Dropout to regularize node embeddings.
    \item “w/ Laplacian”: This variant builds an adjacency matrix based on node distances and applies graph Laplacian regularization to node embeddings.
    \item “w/ Add”: This variant replaces the subtraction operation in RDM with addition.
    \item “w/ Concat”: This variant replaces the subtraction operation in RDM with concatenation.
    \item “w/ $A_{geo}$”: This variant replaces the adjacency matrix in ECO with an adjacency matrix built based on node distances.
    \item “w/ $A_{softmax}$”: This variant replaces the adjacency matrix in ECO with an adaptive adjacency matrix of softmax activation.
\end{itemize}

\begin{table*}[h]
\caption{Performance comparison of different RAGC variants, reporting the average prediction results across 12 time steps.}
\label{tab:ablation-performance}
\centering
\begin{tabular}{|l|ccc|ccc|ccc|ccc|}
\hline
 \multirow{2}{*}{Methods} & \multicolumn{3}{c|}{SD} & \multicolumn{3}{c|}{GBA} & \multicolumn{3}{c|}{GLA} & \multicolumn{3}{c|}{CA} \\ 
 \cline{2-13}
  & MAE & RMSE & MAPE & MAE & RMSE & MAPE & MAE & RMSE & MAPE & MAE & RMSE & MAPE \\ \hline
w/o SSE & 17.32 & 30.60 & 11.17\% & 19.42 & 32.97 & 13.42\% & 19.10 & 33.32 & 10.95\% & 17.45 & 29.95 & 12.08\% \\
w/o RDM & 17.06 & 28.46 & 11.16\% & 22.15 & 33.13 & 14.55\% & 21.27 & 33.92 & 13.93\% & 19.64 & 32.09 & 14.64\% \\
w/o AGC & 17.45 & 30.01 & 11.09\% & 19.47 & 34.07 & 13.51\% & 18.68 & 32.23 & 10.86\% & 17.61 & 30.64 & 12.13\% \\
\hline
w/ Dropout & 17.30 & 29.16 & 11.06\% & 19.34 & 32.95 & 13.08\% & 18.43 & 30.95 & 10.55\% & 17.31 & 29.39 & 11.53\% \\
w/ Laplacian & 17.03 & 29.25 & 11.05\% & 19.10 & 32.58 & 13.25\% & 18.47 & 32.00 & 10.51\% & 17.71 & 31.72 & 11.90\% \\
\hline
w/ Add & 16.35 & 27.83 & 10.71\% & 18.43 & 31.75 & 12.37\% & 18.10 & 30.47 & 10.42\% & 16.54 & 28.41 & 11.02\% \\
w/ Concat & 16.38 & 27.95 & 10.78\% & 18.51 & 31.87 & 12.34\% & 18.09 & 30.37 & 10.46\% & 16.71 & 28.58 & 11.16\% \\
\hline
RAGC & \textbf{16.16} & \textbf{27.40} & \textbf{10.62\%} & \textbf{18.33} & \textbf{31.65} & \textbf{12.18\%} & \textbf{17.75} & \textbf{30.11} & \textbf{10.20\%} & \textbf{16.40} & \textbf{28.23} & \textbf{10.96\%} \\ \hline
\end{tabular}
\end{table*}

\begin{table*}[tp!]
\centering
\caption{Performance comparison of different graph convolution methods.}
\label{tab:ablation-efficiency}
\begin{threeparttable}
\begin{tabular}{|l|ccc|ccc|ccc|ccc|}
\hline
\multirow{2}{*}{Methods} & 
\multicolumn{3}{c|}{SD} & \multicolumn{3}{c|}{GBA} & \multicolumn{3}{c|}{GLA} & \multicolumn{3}{c|}{CA} \\
\cline{2-13}
 & Train & Infer & MAE & Train & Infer & MAE & Train & Infer & MAE & Train & Infer & MAE \\ \hline
w/ $A_{geo}$ & \textbf{13.2} & \textbf{3.8} & 17.06 & \underline{96.7} & \underline{16.2} & 19.49 & \underline{127.7} & \underline{22.2} & 18.45 & \underline{393.1} & \underline{65.5} & 17.43 \\
w/ $A_{softmax}$ & 17.6 & 5.2 & \textbf{16.15} & 103.2 & 16.3 & \underline{18.48} & 153.3 & 23.5 & \underline{17.89} & 525.9 & 67.3 & \underline{16.47} \\
\hline
RAGC & \underline{14.0} & \underline{3.9} & \underline{16.16} & \textbf{65.1} & \textbf{11.5} & \textbf{18.33} & \textbf{92.8} & \textbf{17.5} & \textbf{17.75} & \textbf{203.9} & \textbf{33.4} & \textbf{16.40} \\
\hline
\end{tabular}
\begin{tablenotes}
        \footnotesize
        \item  The best result is in bold, and the second best result is underlined.
      \end{tablenotes}
  \end{threeparttable}
\end{table*}

\begin{figure*}[ht]
  \centering
  \includegraphics[width=0.93\textwidth]{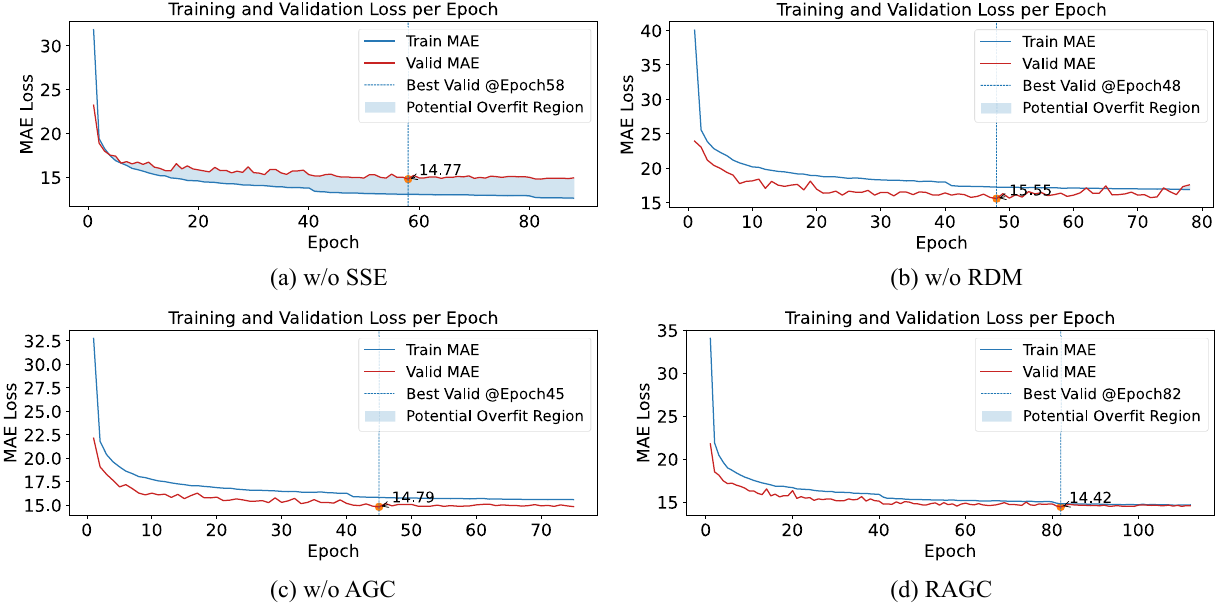}
  \caption{Training and validation loss curves of different RAGC variants on the SD dataset.}
  \label{fig:loss_curve}
\end{figure*}

Tables \ref{tab:ablation-performance} and \ref{tab:ablation-efficiency} present the comparison results across all datasets. First, as shown in Table \ref{tab:ablation-performance}, the ablation variants “w/o SSE”, “w/o RDM”, and “w/o AGC” exhibit a significant decrease in accuracy compared to the full RAGC model. Figure~\ref{fig:loss_curve} further illustrates the training and validation loss curves of the three variants on the SD dataset. In the “w/o SSE” variant, an obvious overfitting region is observed due to the lack of regularization on the node embeddings. In the “w/o RDM” variant, the minimum validation loss remains high at 15.55, accompanied by significant fluctuations in the validation curve, indicating that the noise introduced by SSE adversely affects training stability. The “w/o AGC” variant similarly exhibits underfitting issues. In addition, we tried to replace the subtraction in the residual difference mechanism with addition (“w/ Add”) and concatenation (“w/ Concat”), and achieved suboptimal results, proving the effectiveness of subtractive fusion. In contrast, the full RAGC model achieves smoother loss curves and a lower final validation loss. These results highlight the synergistic effect of SSE, the residual difference strategy, and adaptive graph learning. Specifically, removing SSE eliminates the regularization of node embeddings, leading to over-parameterization; without RDM, the noise introduced by SSE cannot be effectively mitigated, thereby compromising training stability; and without AGC, the aggregation operation along the node dimension is absent, meaning that even with residual difference modeling, the noise from SSE cannot be suppressed effectively.\par
Second, alternative regularization methods such as Dropout and Laplacian regularization fail to achieve satisfactory results. Dropout primarily targets the activation layers rather than the embedding layers, which disrupts the integrity of spatial information. Meanwhile, Laplacian regularization penalizes embedding distances based on a predefined prior graph, lacking data-driven flexibility and being susceptible to noise edges inherent in the prior graph.\par

\begin{table*}[ht]
\centering
\caption{Efficiency study on four large-scale traffic forecasting datasets.}
\label{tab:efficiency}
\begin{threeparttable}
\begin{tabular}{|l|cccc|cccc|}
\hline
\multirow{2}{*}{Methods} & 
\multicolumn{4}{c|}{Training Time (s/epoch)} & \multicolumn{4}{c|}{Inference Time (s/epoch)} \\
\cline{2-9}
 & SD & GBA & GLA & CA & SD & GBA & GLA & CA \\ \hline
GWNet & 154.2 & 375.1 & 657.9 & 2121.5 & 34.1 & 51.4 & 79.8 & 197.0 \\
AGCRN & 72.8 & 803.2 & - & - & 12.1 & 124.9 & - & - \\
DSTAGNN & 265.7 & 2676.2 & - & - & 21.9 & 326.4 & - & - \\
BigST & 27.0 & 80.6 & 131.0 & 286.4 & 4.3 & \underline{7.9} & \underline{11.5} & \underline{23.4} \\
GSNet & \textbf{5.9} & \textbf{17.5} & \textbf{30.3} & \textbf{61.8} & \textbf{1.1} & \textbf{2.6} & \textbf{4.6} & \textbf{10.6} \\
STWave & 231.6 & 796.0 & 1304.9 & 3072.8 & 40.2 & 128.9 & 212.1 & 496.8 \\
PatchSTG & 54.0 & 234.5 & 188.7 & 634.2 & 9.3 & 34.1 & 28.3 & 85.1 \\
\hline
RAGC & \underline{14.0} & \underline{65.1} & \underline{92.8} & \underline{203.9} & \underline{3.9} & 11.5 & 17.5 & 33.4 \\ 
\hline
\end{tabular}
 \begin{tablenotes}
        \footnotesize
        \item  Missing values indicate that the model encountered an out-of-memory error.
        \item  The best result is in bold, and the second best result is underlined.
      \end{tablenotes}
  \end{threeparttable}
\end{table*}

Third, as shown in Table \ref{tab:ablation-efficiency}, we validate the effectiveness of the proposed ECO module by comparing the training time, inference time per epoch, and MAE of different graph convolution methods. “Train” denotes the training time (in seconds) per epoch, “Infer” denotes the inference time (in seconds) per epoch, and “MAE” represents the average prediction results across 12 time steps. It can be observed that using $A_{geo}$, the distance-based predefined adjacency matrix, underperforms compared to using $A_{softmax}$ and the full RAGC model in terms of prediction accuracy, as it cannot capture latent spatial relationships. Regarding computational efficiency, except for the SD dataset (716 nodes), RAGC consistently achieves the fastest training and inference speeds. On smaller datasets, computational efficiency bottlenecks are primarily determined by feature dimension rather than the number of nodes. However, as the number of nodes increases (e.g., in GLA and CA), the linear complexity advantage of ECO becomes increasingly apparent. In summary, ECO ensures high computational efficiency without sacrificing prediction accuracy, making it well-suited for large-scale traffic prediction tasks.

\subsection{Efficiency Comparison}
We compare the training and inference speeds per epoch of RAGC against several baseline models across all four datasets. The baselines include GWNet, AGCRN, and DSTAGNN, which employ graph operations with quadratic complexity, as well as BigST, GSNet, STWave, and PatchSTG, which are designed with computational efficiency optimizations. The comparison results are illustrated in Table \ref{tab:efficiency}. RAGC achieves the second fastest training speed among all methods and ranks third in inference speed. In contrast, models with quadratic complexity (e.g., AGCRN and DSTAGNN) exhibit poor scalability on large-scale datasets (especially on GLA and CA), where memory and computational limitations severely hinder efficiency. GSNet achieves the fastest training and inference speed by compressing spatial adjacency relations. However, this compression results in incomplete modeling of spatial dependencies, which degrades its prediction accuracy. For instance, as shown in Table~\ref{tab:benchmark}, GSNet ranks approximately 10th on the SD dataset when evaluated across all metrics in ascending order. In contrast, apart from GSNet which has poor performace, RAGC almost has the best efficiency. The backbone of RAGC consists solely of multilayer perceptrons and adaptive graph convolution operations, avoiding the use of complex auxiliary modules. Furthermore, by employing a cosine similarity-based adaptive graph without approximation, the proposed ECO module effectively captures the latent spatial dependencies while maintaining linear time complexity. This design allows RAGC to efficiently and accurately model traffic dynamics across large-scale road networks.

\begin{figure}[ht]
  \centering
  \includegraphics[width=0.45\textwidth]{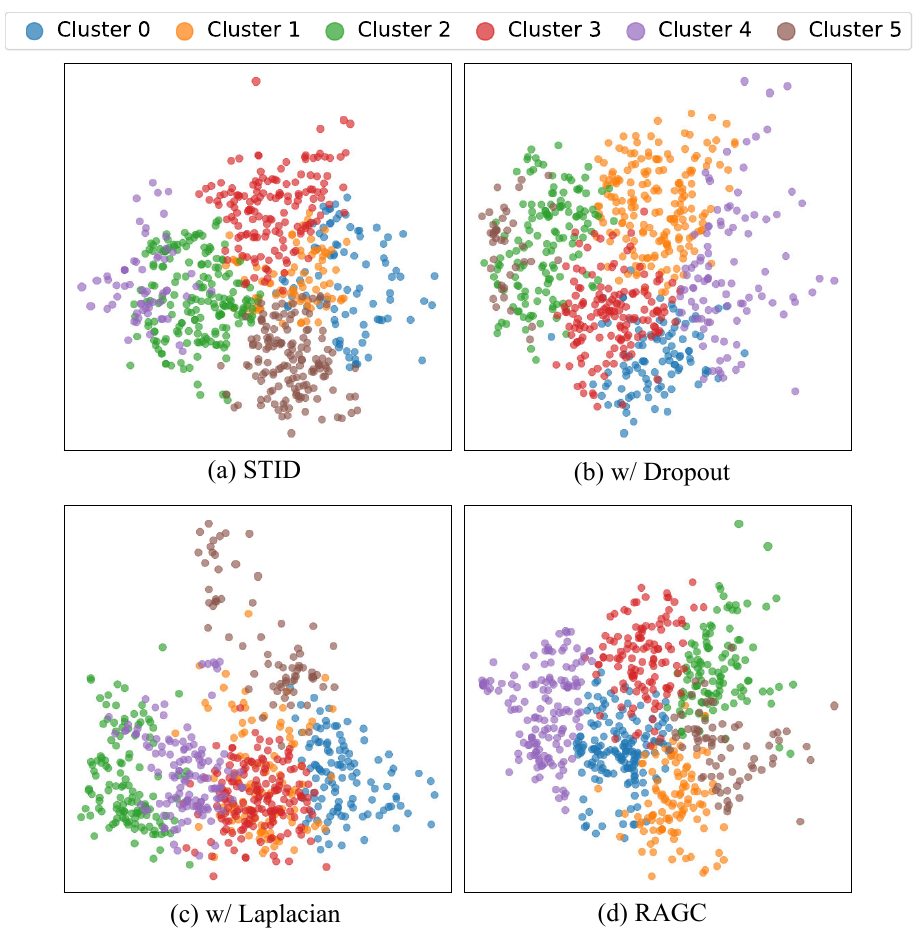}
  \caption{Visualization of node embeddings on the SD dataset after K-Means clustering and PCA dimensionality reduction.}
  \label{fig:vis_node_emb}
\end{figure}

\subsection{Visualization}
\subsubsection{Visualization of Node Embeddings}

\begin{figure*}[ht]
  \centering
  \includegraphics[width=0.9\textwidth]{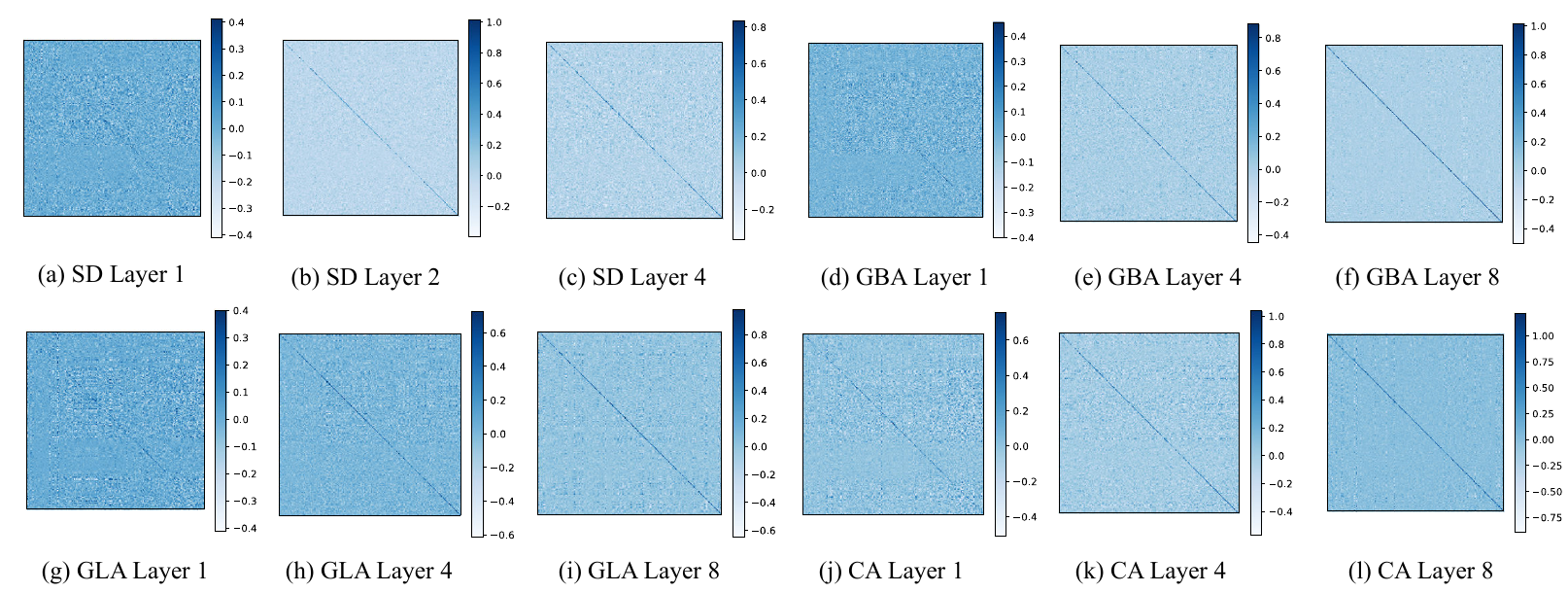}
  \caption{Visualization of $\tilde{W_{g}}$ on four datasets.}
  \label{fig:weights}
\end{figure*}

\begin{figure}[ht]
  \centering
  \includegraphics[width=0.45\textwidth]{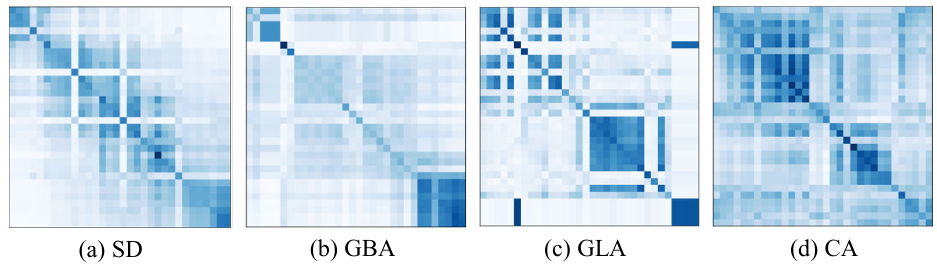}
  \caption{Visualization of learned similarity adjacency matrix on all datasets. For ease of observation, only part of the semantic adjacency matrix is shown.}
  \label{fig:vis_sem_adj}
\end{figure}

\begin{figure}[h!]
  \centering
  \includegraphics[width=0.9\linewidth]{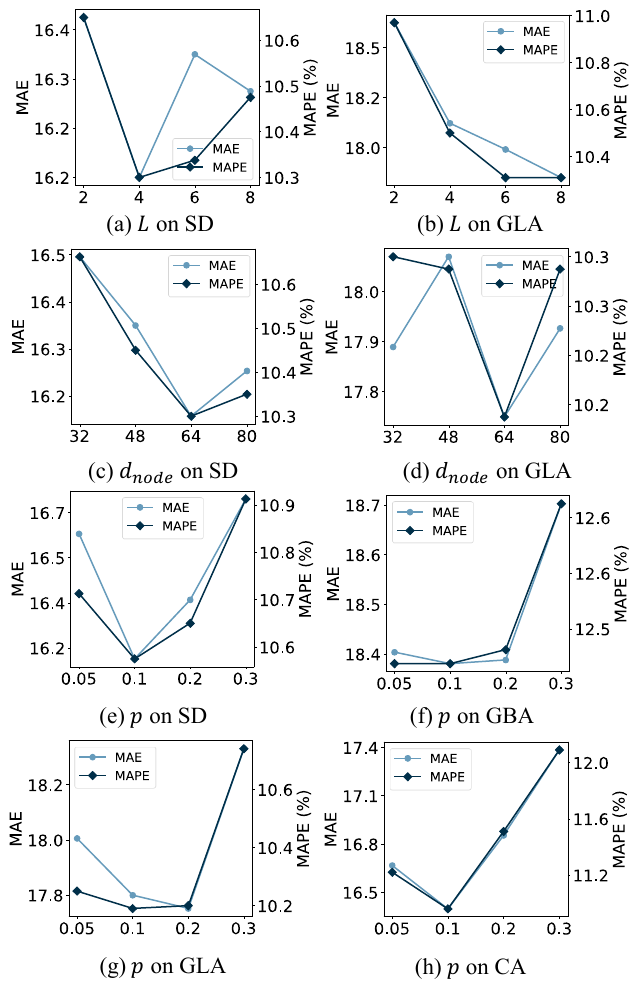}
  \caption{Hyper-parameter study of RAGC.}
  \label{fig:hyperparam}
\end{figure}

We apply K-Means clustering with six clusters to the node embeddings learned by each model and further perform dimensionality reduction using Principal Component Analysis (PCA). Figure~\ref{fig:vis_node_emb} presents the visualization of node embeddings from STID, Dropout, Laplace regularization, and RAGC on the SD dataset, where different node colors indicate different cluster assignments. As shown in the figure, STID does not address the regularization of node embeddings, leading to noticeable overlap between some clusters, such as the green cluster 2 and the purple cluster 4. Dropout randomly zeros out portions of node embeddings during training, disrupting the integrity of spatial information and producing loose clusters, as illustrated in Figure~\ref{fig:vis_node_emb}(b). Laplace regularization, in contrast, relies entirely on the prior graph structure and lacks data-driven adaptability, making it vulnerable to noise in the original graph and causing significant confusion between the orange cluster 1 and the red cluster 3. In contrast, RAGC maintains the flexibility of data-driven embedding regularization, yielding clearer cluster separation and thereby supporting stronger prediction performance.

\subsubsection{Visualization of Weight Matrix}
In Figure~\ref{fig:weights}, we visualize the sum of the learnable weights $W_{g}^{(z)}$ across $Z$ diffusion steps in the adaptive graph convolution, denoted as $\tilde{W_{g}}$. This visualization helps to illustrate the effectiveness of the RDM. In the shallow layers of the model, $\tilde{W_{g}}$ does not closely resemble the identity matrix. This behavior indicates that the model intentionally retains the global average noise term $\bar{e}$ to enhance the regularization effect, as discussed in Equation~\ref{equ_analysis_diff}. In contrast, in the middle and deeper layers, $\tilde{W_{g}}$ progressively approximates the identity matrix. This suggests that the model adaptively suppresses noise to prevent its negative impact on the final time series prediction. Through this adaptive noise-blocking mechanism, the model effectively balances regularization with the preservation of temporal forecasting accuracy.

\subsubsection{Visualization of Similarity Matrix}
We present the cosine similarity adjacency matrices learned by RAGC across all datasets in Figure~\ref{fig:vis_sem_adj}. For clarity, only a portion of each matrix is displayed. As shown, the learned adjacency matrices effectively capture the adjacency relationships between nodes in different datasets.

\subsection{Hyper-parameter Study}
Figure~\ref{fig:hyperparam} presents the search results for hyper-parameters $L$ and $d_{node}$ on the SD and GLA datasets. We evaluate the number of encoder layers $L$ from $\{2, 4, 6, 8\}$ and the node embedding dimension $d_{node}$ from $\{32, 48, 64, 80\}$. As shown in the figure, increasing the number of encoder layers generally improves model performance. For the smaller SD dataset, the best performance is achieved at $L=4$, while for the larger GLA dataset, the best performance occurs at $L=8$. For the node embedding dimension, both datasets achieved the best performance when $d_{node}=64$.\par
Figure~\ref{fig:hyperparam} also shows the search results for the hyper-parameter $p$ across all four datasets. On the GLA dataset, optimal performance is observed at $p = 0.2$, whereas on the SD, GBA, and CA datasets, the best results are achieved when $p = 0.1$. These findings suggest that a replacement probability of 0.1 generally offers a good balance between regularization and model capacity across different scenarios. In contrast, setting $p$ too high may lead to overly strong regularization, which can hinder the model's ability to fit the training data, resulting in underfitting.

\section{Related Work}\label{sec:related_work}

Traffic forecasting is a classic multivariate time series task with significant practical applications. In recent years, deep learning has become the mainstream approach in this field. Early researchers employed Convolutional Neural Networks (CNN) for traffic forecasting \cite{DeepST, ST-ResNet}. However, given the inherent graph structure of traffic networks, Graph Convolutional Networks (GCN) have demonstrated outstanding performance in traffic forecasting \cite{STGCN, GWNet, AGCRN, STFGNN, LSGCN, DyHSL}. Researchers typically use GCNs to capture spatial dependencies among nodes and integrate them with sequence models such as Recurrent Neural Networks (RNN) \cite{DCRNN, AGCRN} and Temporal Convolutional Networks (TCN) \cite{GWNet, STPGNN} to capture temporal dependencies. These combined models are often referred to as Spatial-Temporal Graph Neural Networks (STGNN).\par
The adjacency matrix is one of the core parameters of STGNNs. Some methods construct a fixed adjacency matrix according to predefined rules, including a geographical graph based on sensor distance \cite{STGCN, DCRNN} and a temporal graph based on the similarity of nodes' historical sequences \cite{STFGNN, STGODE}. However, since these adjacency matrices remain fixed throughout the training process, these predefined rules may not fully capture the true spatial dependencies between nodes. \par
An emerging research trend is adaptive graph learning, which employs learnable node embeddings to represent the spatial dependencies among road network nodes. For instance, models such as GWNet \cite{GWNet}, AGCRN \cite{AGCRN}, MTGNN \cite{MTGNN}, and etc. \cite{D2STGNN, HGCN, STG-NCDE, AGS, MegaCRN, HimNet, TGCRN} introduce learnable node embeddings, construct adaptive adjacency matrices for graph convolution, and automatically learn hidden spatial dependencies in a data-driven manner. In contrast, STID \cite{STID} does not explicitly use graph convolution to aggregate spatial information, but instead adopts learnable node embeddings as additional input features. With only simple Multi-Layer Perceptrons (MLP), STID achieves performance comparable to that of STGNNs, demonstrating the potential of node embedding learning for modeling spatial dependencies. This paradigm of using node embeddings as supplementary features has also been widely adopted in subsequent works \cite{STAEFormer, PDFormer, BigST, PatchSTG}. However, node embeddings often account for a large proportion of the model parameters, which can easily lead to over-parameterization and thus limit the effectiveness of adaptive graph learning.
Several studies have explored regularization techniques for ST-GCNs, such as RGSL \cite{RGSL}, which is based on sparse adjacency matrices, and STC-Dropout \cite{STC-Dropout}, which leverages curriculum learning and dropout. However, these methods regularize the adaptive adjacency matrix or graph signal and do not directly process node embeddings, thereby failing to fundamentally solve the overfitting problem associated with node embeddings and complicating the training or inference process. A novel regularization technique named Stochastic Shared Embeddings (SSE) has been proposed for recommendation systems \cite{SSE, SSE-PT}. SSE regularizes the embedding layer by randomly replacing the embeddings of different nodes during training, thereby introducing diversity and helping to explore richer latent semantic relationships. However, in the context of traffic flow prediction, where high continuity and precision of the input signal are paramount, the random replacement of adjacent node embeddings may introduce noise that distorts the spatial-temporal signals. Consequently, the model might capture incorrect spatial-temporal dependencies, and these errors could compound over multiple layers, ultimately undermining training stability. In this paper, we combine both SSE and adaptive graph convolution to efficiently capture complex spatial-temporal dependencies.\par

\section{Conclusion}\label{sec:conclusion}
In this paper, we propose a Regularized Adaptive Graph Convolution (RAGC) model for traffic prediction. We find that existing adaptive graph learning methods suffer from $O(N^2)$ computational complexity and lack regularization for node embeddings, which constitute a significant portion of the model parameters. To address these issues, RAGC introduces Efficient Convolution Operator (ECO) to reduces the computational complexity of adaptive graph convolution to $O(N)$, ensuring scalability to large-scale road networks. Furthermore, RAGC employs Stochastic Shared Embedding (SSE) to regularize node embeddings and integrates a residual difference mechanism with adaptive graph convolution to block noise propagation. This design yields a synergistic effect among SSE, the residual difference mechanism, and adaptive graph learning. Extensive experiments on four large-scale benchmark datasets demonstrate that RAGC consistently outperforms state-of-the-art methods in prediction accuracy while maintaining competitive computational efficiency.

\section*{Acknowledgments}The authors would like to thank the anonymous reviewers for their helpful comments. This work was supported by the NSFC 61572537, and the CCF-Huawei Populus Grove Challenge Fund 202305. Yubao Liu is the corresponding author.

\section*{AI-Generated Content Acknowledgement}Large Language Models (LLMs) were used to assist in the writing and polishing of this manuscript. Specifically, we employed an LLM to refine the language, improve readability, and enhance clarity in various sections of the paper. The assistance included sentence rephrasing, grammar checking, and improving the overall flow of the text. Importantly, the LLM was not involved in the conception of the study, research methodology, experimental design, or data analysis. All research ideas, technical contributions, and analytical processes were independently developed and conducted by the authors. The LLM’s role was limited strictly to improving the linguistic quality of the manuscript, without influencing any scientific content. The authors take full responsibility for the content of this manuscript, including any text that was generated or polished with the help of the LLM.

\vspace{12pt}

\end{document}